\documentclass[runningheads]{llncs}

\usepackage[T1]{fontenc}
\usepackage{graphicx}
\usepackage{amsmath,amssymb}
\usepackage{booktabs}
\usepackage{hyperref}
\usepackage{cite}
\usepackage{multirow}
\usepackage{makecell}
\usepackage{array}
\usepackage{siunitx}

\begin{document}

% -------- TITLE & AUTHORS (ANONYMIZED) --------
\title{Attention-Gated Convolutional Networks for Scanner-Agnostic Quality Assessment}

\titlerunning{Attention-Gated CNNs for Quality Assessment}

% For anonymized submission, keep authors and affiliations generic as per MICCAI rules.
\author{Chinmay Bakhale \hspace{1cm}
Anil Kumar Sao}
\institute{Indian Institute of Technology, Bhilai, India \\
\email{}}

\maketitle

% -------- ABSTRACT & KEYWORDS --------
\begin{abstract}
Motion artifacts present a significant challenge in structural MRI (sMRI), often compromising clinical diagnostics and large-scale automated analysis. While manual quality control (QC) remains the gold standard, it is increasingly unscalable for massive longitudinal studies. To address this, we propose a hybrid CNN-Attention framework designed for robust, site-invariant MRI quality assessment. Our architecture integrates a hierarchical 2D CNN encoder for local spatial feature extraction with a multi-head cross-attention mechanism to model global dependencies. This synergy enables the model to prioritize motion relevant artifact signatures, such as ringing and blurring, while dynamically filtering out site-specific intensity variations and background noise. The framework was trained end-to-end on the MR-ART dataset using a balanced cohort of 200 subjects. Performance was evaluated across two tiers: Seen Site Evaluation on a held-out MR-ART partition and Unseen Site Evaluation using 200 subjects from 17 heterogeneous sites in the ABIDE archive. On seen sites, the model achieved a scan-level accuracy of 0.9920 and an F1-score of 0.9919. Crucially, it maintained strong generalization across unseen ABIDE sites (Acc = 0.755) without any retraining or fine-tuning, demonstrating high resilience to domain shift. These results indicate that attention-based feature re-weighting successfully captures universal artifact descriptors, bridging the performance gap between diverse imaging environments and scanner manufacturers.
\end{abstract}

% -------- INTRODUCTION --------
% TODO:
% - Introduce the clinical/technical problem and its importance.
% - Summarize limitations of existing approaches with key references.
% - Clearly state the gap your work addresses.
% - List your main contributions in 2--4 sentences (e.g., "Our main contributions are: ...").
% - Optionally outline the structure of the paper.

% Example contribution paragraph:
% Our main contributions are: (1) <contribution 1>, (2) <contribution 2>, and (3) <contribution 3>.
\section{Introduction}
Automated structural MRI (sMRI) analysis is heavily dependent on raw scan quality~\cite{backhausen2016quality, reuter2015head}. Motion artifacts are a primary concern in clinical cohorts such as Autism Spectrum Disorder (ASD), Attention-Deficit/Hyperactivity Disorder (ADHD) where data acquisition is demanding and rigorous Quality Control (QC) is essential for valid conclusions~\cite{di2014autism, nordahl2016methods,reuter2015head,rauch2005neuroimaging,Naveethaarxiv}. While manual QC is feasible for small studies, it is not easily scalable to large longitudinal efforts like the  Adolescent Brain Cognitive Development (ABCD) study, or  Alzheimer's Disease Neuroimaging Initiative (ADNI)~\cite{casey2018adolescentabcd, miller2016multimodal,jack2008alzheimer}. While literature extensively explores retrospective and prospective MRI quality enhancement, these methods see limited clinical adoption. This stagnation stems from a dual challenge: institutional resistance to modifying established acquisition protocols and the poor generalization of current AI models across diverse imaging environments~\cite{eichhorn2021characterisation,kaurimaging,Kaur2023_Estimation3TMR,Kecskemeti2018_RobustMotionCorrection}.

Conventional approaches typically focus on estimating features using Image Quality Metrics (IQMs) followed by classification using machine learning (ML) methods like Support Vector Machines (SVMs)~\cite{esteban2017mriqc, white2018automated}, but these are often computationally expensive due to required preprocessing like skull-stripping and tissue segmentation. While Convolutional Neural Networks (CNNs) accelerate this by eliminating ad hoc feature engineering~\cite{sujit2019automated}, recent findings show that both deep learning (DL) and traditional ML perform poorly on ``unseen" sites—scanners not included in the training set~\cite{kaur2024quality, Prabhjot,vakli2023automatic}. Even interpretable models like Prototypical Networks though attempts much needed display of showing slice region with artifact yet struggle with generalization to new scanner data~\cite{garcia2024brainqcnet}.

This represents a critical gap: real-world scalability is limited by poor generalization to new imaging environments. Recent efforts have attempted to bridge this via advanced training strategies, such as nested cross-validation for hyperparameter optimization~\cite{bhalerao2025automated} or histogram-based features to combat site-wise variability~\cite{Naveetha}. However, these methods still depend on complex, multi-stage pipelines or handcrafted selections. These challenges are being addressed in this work from a purely architectural perspective, investigating whether the nuances of multi-step QC can be automated within the model itself using attention modules to isolate site-invariant artifact signatures.

CNNs dominate MRI analysis due to their spatial inductive bias but struggle to capture the global dependencies required for complex anatomical reasoning~\cite{lecun1998gradient, krizhevsky2012imagenet}. Conversely, Vision Transformers (ViT) excel at capturing long-range relationships~\cite{dosovitskiy2020image} but demand massive data for training which is often impractical for medical quality assessment (QA). Hybrid CNN-Attention models are not new, and provide a ``best of both worlds'' a proven solution for generalization~\cite{dai2021coatnet, chen2021transunet}; yet, this synergy remains unexplored for site-invariant MRI QA.

We hypothesize that ``marrying" local CNN feature extraction with Attention-based global weighting eliminates the necessity for labor-intensive multi-stage training and handcrafted feature engineering. This hybrid approach provides superior robustness without the prohibitive data requirements of pure Transformers. By integrating attention modules to represent convolutional features prior to classification, we trained our model on the MR-ART dataset and evaluated its performance on both unseen MR-ART/ABIDE subjects. By evaluating our model across 17 heterogeneous sites from the ABIDE archive, we address the critical gap in unseen-site generalization that has historically hindered automated quality control. This work represents the first application of attention mechanisms to MRI quality assessment, demonstrating a superior ability to bridge the performance gap on unseen sites by simultaneously maximizing usable scan retention and maintaining high sensitivity to subtle, site-specific artifacts. 

\section{Method}

\subsection{Dataset}
We utilize the MR-ART dataset for training and Seen Site Evaluation. It contains T1-weighted scans of subjects under three conditions: one motion-free reference and two with varying instructed motion. This paired design isolates motion signatures from anatomical variance, with ground truth provided by expert manual labels. For Unseen Site Evaluation, we use the ABIDE archive, sampling 200 subjects across 17 heterogeneous sites. This diverse cohort—spanning multiple scanner manufacturers and field strengths—serves as a rigorous test for site-invariant generalization.

Volumetric data are processed as 2D axial slices. To focus on informative anatomy, we extract the middle 50 slices per volume and discard those with mean intensity $<0.01$ to remove any background noise. Each slice undergoes independent min-max normalization to $[0, 1]$. This per-slice scaling is critical for the cross-attention mechanism, as it forces the model to prioritize structural textures and artifact patterns over global signal intensity variations across different sites.

\subsection{Problem Formulation}
We define the detection of motion artifacts as a binary classification task. Given a 2D axial slice $x \in \mathbb{R}^{H \times W}$ extracted from a 3D MRI volume, we seek to learn a mapping $f_\theta: x \rightarrow \hat{y}$, where $\hat{y} \in [0, 1]$ represents the probability that $x$ is motion-corrupted ($y=1$) versus clean ($y=0$). The hyper parameters $\bf{\theta}$ are optimized by minimizing the Binary Cross Entropy (BCE) loss:\begin{equation}\mathcal{L}(\theta) = -\frac{1}{N} \sum_{i=1}^{N} \left[ y_i \log(f_\theta(x_i)) + (1 - y_i) \log(1 - f_\theta(x_i)) \right],\end{equation}where $N$ denotes the batch size. This objective ensures that $f_\theta$ produces well-calibrated probability estimates by penalizing structural deviations from the expert-annotated ground truth.

\subsection{Proposed Approach}

\begin{figure}[t]
    \centering
    \includegraphics[width=1.0\textwidth]{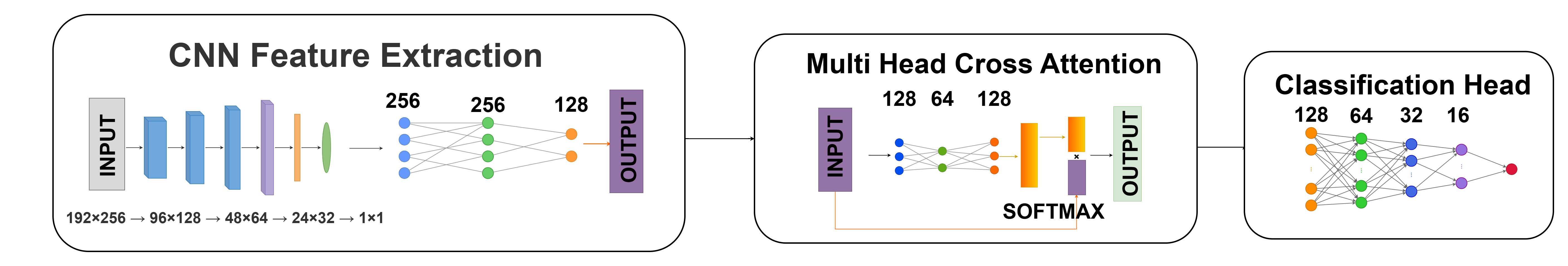}
    \caption{Proposed model architecture comprising a CNN feature extractor (192×256 → 1×1), a self-attention module (128→64→128), and a classification head (128→64→32→16→1) for MRI motion artifact detection.}
    \label{fig:arch}
\end{figure}

The proposed framework comprises three sequential stages, as illustrated in Fig \ref{fig:arch}, embedded in an end-to-end model:(i) a 2D CNN encoder for local spatial feature extraction, (ii) a multi-head attention suite for global feature re-weighting, and (iii) a classification head for quality prediction. The CNN captures slice-level morphological textures with spatial correlations often relevant to the task, while the attention mechanism models long-range dependencies to prioritize task-relevant artifact signatures over site-specific noise. This decoupling enables the model to map diverse imaging inputs into a single, site-invariant latent space for binary quality categorization.

\subsubsection{Feature Extraction using CNN}
The local feature extraction branch consists of a deep CNN organized into three residual-style blocks, totaling six convolutional layers. Each block employs two $3 \times 3$ convolutional layers with zero-padding to preserve spatial resolution, followed by batch normalization and Rectified Linear Unit (ReLU) activations to stabilize training. Spatial downsampling is achieved via max-pooling layers (stride 2) after the first two blocks, while the final block utilizes an adaptive average pooling layer to ensure a dimension-invariant latent representation. To facilitate high-level feature encoding, the channel depth is progressively expanded to 256. This hierarchical architecture is specifically designed to capture fundamental morphological cues—such as edge discontinuities, fine textures, and localized intensity fluctuations—that serve as primary indicators of motion-induced artifacts in structural MRI. The pooled features are subsequently projected into 128-dimensional embedding vectors to facilitate downstream attention based processing. 

\subsubsection{Attention Modules}
To synthesize the extracted features, we employ a cross-attention mechanism that models inter-dependencies within the latent space. The 128-dimensional CNN embeddings are processed through a two-layer attention network with a hidden dimensionality of 256. {We employ channel-wise spatial self-attention, where attention is applied independently to each feature channel over its flattened spatial dimensions, analogous to depthwise convolution but with content-dependent weights.} By applying a softmax activation over the feature distribution, the module generates weights that dynamically redistribute importance across the feature space. This attention mechanism is critical for site-invariance; it serves as a global reasoning layer that evaluates the ``importance'' of convolutional features relative to the entire image. By selectively weighting these features, the model learns to prioritize task-relevant artifact signatures—such as ringing or blurring—while actively suppressing low-variance, site-specific background noise that often leads to poor generalization in pure CNN architectures. 

\subsubsection{Classification Head - Fully Connected Layers} The attended representations are fed into a multi-layer perceptron (MLP) classification head consisting of fully connected layers with decreasing dimensionality ($256 \rightarrow 128 \rightarrow 64 \rightarrow 1$). To combat the ``domain shift" encountered when moving from MR-ART training data to the 17 heterogeneous ABIDE sites, we utilize dropout regularization to prevent the model from over-relying on any single feature dimension. This ensures the classifier focuses on robust, generalized quality indicators. A final sigmoid activation yields a scalar probability, $p \in [0, 1]$, providing a continuous metric for automated scan rejection that can be tuned for varying sensitivity requirements across different clinical cohorts.

\subsubsection{Model Training}
The proposed architecture—comprising the CNN encoder, cross-attention modules, and fully connected classification head—was trained in an end-to-end fashion. This ensures that the convolutional filters and attention weights are jointly optimized to extract and prioritize features specifically relevant to artifact detection. The training cohort consisted of a balanced set of 100 subjects from the MR-ART dataset (50 motion-free score 1 and 50 motion-corrupted score 3 scans). We utilized the Adam optimizer with an initial learning rate of $10^{-3}$ and a batch size of 16. To ensure convergence, we implemented a learning rate scheduler that reduced the rate by a factor of 0.5 upon a plateau in validation loss with a patience of 20 epochs. To stabilize the training of the hybrid layers and prevent gradient explosion, gradient clipping with a maximum norm of 1.0 was applied. Training was conducted for 50 epochs, incorporating an early stopping mechanism (20 epochs) that monitored validation loss and restored the optimal weights to mitigate overfitting. All experiments were implemented in PyTorch and executed using CUDA-accelerated hardware.

\subsection{Model Evaluation and Inference}
The model is trained to predict the quality of input scan as good or bad formulated as binary classification. Since the input to model is 2D slice, to transition from slice-wise predictions to a holistic scan-level assessment, we employ a majority voting aggregation strategy. A scan is categorized as "Poor Quality" (motion-corrupted) if more than 50\% of its constituent slices are classified as such. The rationale for this threshold is to account for the inherent ambiguity in individual slices; while localized noise may occasionally trigger false positives, the aggregate pattern across the volume provides a more robust indicator of systemic motion. Subject identities are tracked during inference to ensure precise grouping of slice-level outputs.

\subsubsection{Evaluation on Seen Site:} 
We established a performance baseline using a held-out partition of the MR-ART dataset, focusing on subject-level generalization within a static imaging environment. The test set comprised 100 subjects, balanced equally between motion-free (score 1; $n=50$) and motion-corrupted (score 3; $n=50$) scans. Evaluating on subjects not encountered during training allows us to verify the model’s fundamental capacity to detect artifact signatures when scanner hardware and acquisition parameters remain constant, isolating architectural efficacy from environmental variability.

\subsubsection{Evaluation on Unseen Sites:}
To assess real-world scalability and robustness against domain shift, we performed extensive testing on the ABIDE dataset across 17 heterogeneous sites. This "unseen site" evaluation is critical, as the model was exposed to acquisition protocols and scanner models entirely absent from the training phase. We utilized a balanced cohort of 200 subjects (100 motion-free; 100 motion-corrupted) sampled across the 17 sites. This rigorous cross-site validation demonstrates the model's ability to extract site-invariant features, proving its utility for harmonizing quality control in large-scale, multi-center neuroimaging consortia. 

% -------- EXPERIMENTS AND RESULTS --------
\section{Experimental Results}
%\subsection{Subject level generalization: Evaluation on MR-ART dataset} 
Table \ref{tab:results} shows performance on the MRART test set. At the slice level, our model achieved 98.6 percent accuracy with 99.6 percent precision and 97.6 percent recall. The AUC ROC score was 0.999. At the scan level using majority voting, accuracy improved to 99.2 percent with perfect precision and 98.4 percent recall.

% \begin{table}[t]
% \centering
% \caption{Performance on MRART Test Set}
% \label{tab:mrart_results}
% \begin{tabular}{@{}lcc@{}}
% \toprule
% Metric & Slice Level & Scan Level \\
% \midrule
% Accuracy & 0.9859 & 0.9920 \\
% Precision & 0.9959 & 1.0000 \\
% Recall & 0.9758 & 0.9840 \\
% F1 Score & 0.9857 & 0.9919 \\
% AUC ROC & 0.9994 & -- \\
% \bottomrule
% \end{tabular}
% \end{table}

%\subsection{Site level Generalization: Evaluation on ABIDE dataset}
Table \ref{tab:results} also illustrates the cross dataset results. It can be observed, despite the differences in scanner manufacturers and acquisition protocols, the model maintained strong performance. %At the slice level, accuracy remained high. At the scan level, the model achieved robust accuracy demonstrating that the learned features generalize well across different imaging sites.

% \begin{table}[t]
% \centering
% \caption{Slice-level and Scan-level Performance on MR-ART and ABIDE}
% \label{tab:results}
% \begin{tabular}{@{}l cc cc@{}}
% \toprule
% & \multicolumn{2}{c}{\textbf{MR-ART}} & \multicolumn{2}{c}{\textbf{ABIDE}} \\
% \cmidrule(lr){2-3} \cmidrule(lr){4-5}
% \textbf{Metric} & \textbf{Slice } & \textbf{Scan } & \textbf{Slice} & \textbf{Scan} \\
% \ & \textbf{ Level} & \textbf{Level} & \textbf{Level} & \textbf{Level} \\
% \midrule
% Accuracy  & 0.9859 & 0.9920 & 0.7384 & 0.7550 \\
% Precision & 0.9959 & 1.0000 & 0.7074 & 0.7249 \\
% Recall    & 0.9758 & 0.9840 & 0.8368 & 0.8460 \\
% F1 Score  & 0.9857 & 0.9919 & --     & --     \\
% AUC-ROC   & 0.9994 & --     & 0.8193 & --     \\
% \bottomrule
% \end{tabular}
% \end{table}

\begin{table}[t]
\centering
\caption{Slice-level and Scan-level Performance on MR-ART and ABIDE}
\label{tab:results}

\setlength{\tabcolsep}{10pt}  % increase horizontal spacing
\renewcommand{\arraystretch}{1.1}

\begin{tabular}{l c c c c}
\toprule
& \multicolumn{2}{c}{\textbf{MR-ART}} & \multicolumn{2}{c}{\textbf{ABIDE}} \\
\cmidrule(lr){2-3} \cmidrule(lr){4-5}
\textbf{Metric} & \textbf{Slice Level} & \textbf{Scan Level} & \textbf{Slice Level} & \textbf{Scan Level} \\
\midrule
Accuracy  & 0.9859 & 0.9920 & 0.7384 & 0.7550 \\
Precision & 0.9959 & 1.0000 & 0.7074 & 0.7249 \\
Recall    & 0.9758 & 0.9840 & 0.8368 & 0.8460 \\
F1 Score  & 0.9857 & 0.9919 & --     & --     \\
AUC-ROC   & 0.9994 & --     & 0.8193 & --     \\
\bottomrule
\end{tabular}
\end{table}

% \begin{table}[t]
% \centering
% \caption{Slice-level and Scan-level Performance on MRART and ABIDE}
% \label{tab:results}
% \begin{tabular}{@{}llcc@{}}
% \toprule
% Dataset & Metric & Slice Level & Scan Level \\
% \midrule
% \multirow{5}{*}{MRART}
%  & Accuracy  & 0.9859 & 0.9920 \\
%  & Precision & 0.9959 & 1.0000 \\
%  & Recall    & 0.9758 & 0.9840 \\
%  & F1 Score  & 0.9857 & 0.9919 \\
%  & AUC-ROC   & 0.9994 & --     \\
% \midrule
% \multirow{4}{*}{ABIDE}
%  & Accuracy  & 0.7384 & 0.7550 \\
%  & Precision & 0.7074 & 0.7249 \\
%  & Recall    & 0.8368 & 0.8460 \\
%  & AUC-ROC   & 0.8193 & --     \\
% \bottomrule
% \end{tabular}
% \end{table}

\begin{figure}[t]
    \centering
    \includegraphics[width=0.8\textwidth]{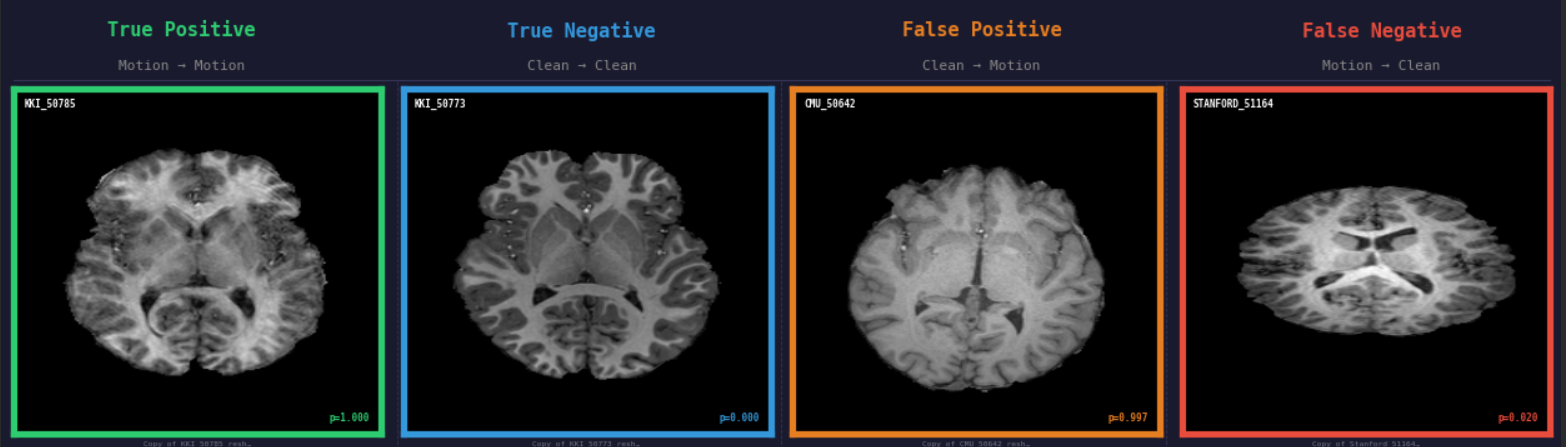}
    \caption{Examples of correctly and incorrectly predicted images from ABIDE dataset.}
    \label{fig:samples1}
\end{figure}

\begin{figure}[t]
    \centering
    \includegraphics[width=0.8\textwidth]{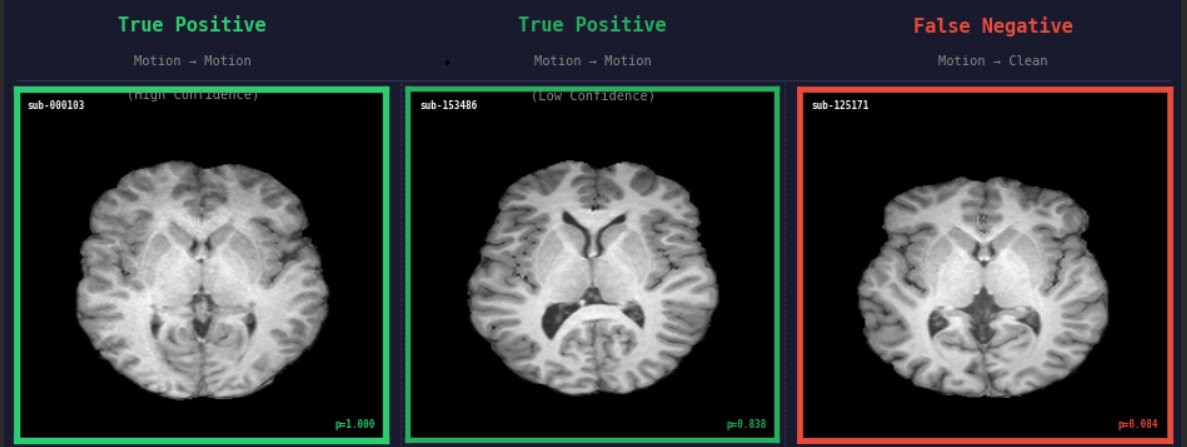}
    \caption{Examples of correctly and incorrectly predicted images from MRART dataset.}
    \label{fig:samples2}
\end{figure}

%\subsection{Qualitative Results }
Figures \ref{fig:samples1} and \ref{fig:samples2} represent examples of correctly and incorrectly predicted images from the MR-ART and ABIDE dataset. We have mentioned the probability of FC layer impacting classificaiton label here. 

%\subsection{Ablation Study}
Table~\ref{tab:ablation_study} presents the ablation study results on both 
MR-ART and ABIDE datasets. On MR-ART, all three configurations perform 
comparably well, with the full model (CNN + Attention + Classification Head) 
achieving the highest scan level accuracy of 0.9920 and a recall of 0.9840, 
confirming that each component contributes to in-distribution performance. 
The more appealing results are on ABIDE, where the model is tested on entirely 
unseen sites. Adding the Classification Head alone sharply reduces recall to 
0.69, suggesting it overfits to MR-ART specific patterns. The full model, 
with both attention and the Classification Head, recovers recall to 0.8460, 
showing that the attention module plays an important role in retaining 
sensitivity when generalizing to new sites, though at a moderate cost to 
accuracy (0.7550).

\begin{table}[ht]
\centering
\caption{%
    Ablation Study Results on MR-ART and ABIDE Datasets.
    Scan-level predictions use majority voting over the middle 50 axial slices.
}
\label{tab:ablation_study}

\setlength{\tabcolsep}{4pt}   % reduced horizontal padding
\renewcommand{\arraystretch}{1.3}

\begin{tabular}{
l 
p{4cm}   % fixed width for configuration (wraps text)
S[table-format=1.3] 
S[table-format=1.3] 
S[table-format=1.3] 
S[table-format=1.3]
}
\toprule    

\multirow{2}{*}{\textbf{Dataset}}
& \multirow{2}{*}{\textbf{Configuration}}
& \multicolumn{2}{c}{\textbf{Scan Level}}
& \multicolumn{2}{c}{\textbf{Slice Level}} \\

\cmidrule(lr){3-4} \cmidrule(lr){5-6}

& & {\textbf{Acc}} & {\textbf{Recall}} & {\textbf{Acc}} & {\textbf{Recall}} \\

\midrule

\multirow{3}{*}{\textbf{MR-ART}}
  & CNN + Attention
  & 0.99 & 0.98 & 0.9818 & 0.9768 \\
  & CNN + Classification Head
  & 0.98 & 0.96 & 0.9684 & 0.9372 \\
  & CNN + Attention + Classification Head
  & 0.9920 & 0.9840 & 0.9859 & 0.9758 \\
\midrule

\multirow{3}{*}{\textbf{ABIDE}}
  & CNN + Attention
  & 0.78 & 0.78 & 0.7761 & 0.7934 \\
  & CNN + Classification Head
  & 0.795 & 0.69 & 0.7879 & 0.6848 \\
  & CNN + Attention + Classification Head
  & 0.7550 & 0.8460 & 0.7384 & 0.8368 \\

\bottomrule
\end{tabular}

\raggedright
\footnotesize
Acc\,=\,Accuracy;\;
Recall\,=\,Sensitivity (True Positive Rate).

\end{table}

%\subsection{Quantitative Results}
Table \ref{tab:comparison} compares the proposed method with three existing approaches for MRI motion artifact classification. On MR-ART, the proposed method achieves 99.20\% accuracy and 98.40\% sensitivity, outperforming all compared methods. The MIA method, despite training on a larger multi-dataset combination, reaches only 84.17\% accuracy on the same test set. When tested on the unseen ABIDE dataset without retraining, the proposed method achieves 75.50\% accuracy and 84.60\% sensitivity, demonstrating reasonable cross-dataset generalisation. Notably, this sensitivity exceeds the JMRI method's 77\%, even though JMRI~\cite{jimeno2024automated} was both trained and tested on ABIDE.

\begin{table}[ht]
\centering
\caption{Comparison of the proposed method with existing literature on MRI motion artifact classification.}
\label{tab:comparison}

\small
\setlength{\tabcolsep}{3pt}   % slightly tighter horizontal padding
\renewcommand{\arraystretch}{1.3}

\begin{tabular}{|p{1.65cm}|p{4.0cm}|p{2.3cm}|p{1.85cm}|p{1.0cm}|p{1.0cm}|}
\hline
\textbf{Literature} & \textbf{Method} & \textbf{Training Dataset} & \textbf{Test Dataset} & \textbf{Acc.} & \textbf{Sens.} \\
\hline

MIA~\cite{vakli2023automatic}
& Full volume fed into 3D-CNN model
& MR-ART, UKBB, OASIS-3, In-house
& MR-ART
& 84.17\%
& 87.91\% \\
\hline

JMRI~\cite{jimeno2024automated}
& Slice-level evaluation per orientation and ensembling for final volume class
& ABIDE
& ABIDE
& 84\%
& 77\% \\
\hline

IJNS~\cite{roecher2024motion}
& Stacks of slices classified and aggregated for final volume class
& 1000BRAINS
& 1000BRAINS
& 95\%
& - \\
\hline

\parbox[t]{1.3cm}{\raggedright Naveetha et al.~\cite{Naveetha}}
& \parbox[t]{4.2cm}{\raggedright DHoGM feature + 4-layer MLP (209 params); majority voting}
& MR-ART & MR-ART & 93.00\% & 90.00\% \\
\cline{3-6}

& & MR-ART & ABIDE  & 88.00\% & 89.00\% \\
\hline

\parbox[t]{1.3cm}{\raggedright Proposed Method}
& \parbox[t]{4.2cm}{\raggedright Per-slice CNN + self-attention reweighting; majority voting}
& MR-ART & MR-ART & 99.20\% & 98.40\% \\
\cline{3-6}

& & MR-ART & ABIDE  & 75.50\% & 84.60\% \\
\hline

\end{tabular}
\end{table}

\section{Conclusion}
We proposed a hybrid CNN-Attention framework for site-invariant MRI quality assessment. By integrating multi-head attention with convolutional encoders, the model filters site-specific intensity variations to isolate universal motion signatures. Our approach achieves high accuracy on seen sites ($Acc = 0.9920$) and generalizes effectively across 17 unseen sites in the ABIDE dataset ($Acc = 0.755$) without retraining, demonstrating the potential of attention-based re-weighting to bridge domain gaps in multi-center consortia. While promising, this study is limited by its current cohort size and scope of architectural comparisons. Future work will evaluate the framework on larger, more diverse datasets—including the full ABIDE ($900+$ scans), ABCD ($2400+$ scans), and ADNI cohorts—and include exhaustive ablation studies to rigorously quantify the performance gains of the hybrid architecture against standalone CNN and attention-only benchmarks. 
% TODO:
% - Summarize the problem tackled and your main contributions.
% - Recap key empirical findings in 2--3 sentences.
% - Mention potential impact or applications.
% - Outline 1--2 directions for future work.

% Example:
% In this work, we proposed <method> for <task>, achieving improved performance over existing approaches on <dataset>. 
% Future work will explore <direction 1> and <direction 2>.

% -------- OPTIONAL ACKNOWLEDGMENTS --------
% \paragraph{Acknowledgments.}
% OPTIONAL for initial submission:
% TODO: Add funding sources, data providers, and collaborators here if allowed.
% Example: This study was funded by <Agency> (grant number <ID>).
\newpage

% -------- OPTIONAL DISCLOSURE OF INTERESTS --------
% \paragraph{Disclosure of Interests.}
% OPTIONAL for initial submission:
% TODO: Either explicitly state no competing interests or disclose them.
% Example: The authors have no competing interests to declare that are relevant to the content of this article.

% -------- REFERENCES -------

\bibliographystyle{splncs04}
\bibliography{ref}
% Create a "references.bib" file with your BibTeX entries.

\end{document}